\documentclass[letterpaper]{article} 
\usepackage{aaai2027}
\nocopyright
\usepackage[hyphens]{url}  
\usepackage{graphicx} 
\usepackage{natbib}  
\usepackage{caption} 
\usepackage{algorithm}
\usepackage{algorithmic}

\usepackage{newfloat}
\usepackage{listings}
\DeclareCaptionStyle{ruled}{labelfont=normalfont,labelsep=colon,strut=off} 
\floatstyle{ruled}
\newfloat{listing}{tb}{lst}{}
\floatname{listing}{Listing}

\usepackage{booktabs}

\usepackage{amsmath}
\usepackage{amsfonts}
\usepackage{xcolor} 
\usepackage{pifont}
\usepackage[table]{xcolor}
\usepackage{multirow}

\title{Ripple: Real-Time Streaming Audio-Video\\ Generation With Cross-Modal Recurrent Memory}
\author{
    Yanbo Ding\textsuperscript{\rm1,2,4,}\thanks{Work done during an internship at China Telecom Artificial Intelligence Technology (Beijing) Co., Ltd.}, 
    Zhizhi Guo\textsuperscript{\rm2,}\thanks{Project leader.},
    Quanyue Song\textsuperscript{\rm2,5,}\footnotemark[1],\\
    Yishan He\textsuperscript{\rm2}, 
    Zhixiang He\textsuperscript{\rm2}, 
    Yongxiang Li\textsuperscript{\rm2}, 
    Yali Wang\textsuperscript{\rm1,3,}\corresponding
}
\affiliations{
    \textsuperscript{\rm 1}Shenzhen Key Lab of Computer Vision and Pattern Recognition,\\ Shenzhen Institutes of Advanced Technology, Chinese Academy of Sciences\\
    \textsuperscript{\rm 2}China Telecom Artificial Intelligence Technology (Beijing) Co., Ltd., China\\
    \textsuperscript{\rm 3}Shanghai Artificial Intelligence Laboratory\\
    \textsuperscript{\rm 4}School of Artificial Intelligence, University of Chinese Academy of Sciences\\
    \textsuperscript{\rm 5}State Key Laboratory of Human-Machine Hybrid Augmented Intelligence,\\Institute of Artificial Intelligence and Robotics, Xi’an Jiaotong University, China
}

\begin{document}

\maketitle

\begin{abstract}
 Audio-video generative models achieve impressive quality but suffer from high latency, making them unsuitable for real-time applications. Although several streaming audio-video generation methods have been proposed, they remain costly and fail to support long-form generation. To address this, we propose \textbf{Ripple}, a real-time joint audio-video generation system with a cross-modal recurrent memory mechanism.
To enable efficient streaming inference while preserving long-term context, Ripple combines a fixed-length sliding-window attention with modality-specific memory states that continuously summarize audio and video context. Cross-modal memory interaction is further introduced to enhance audio-visual synchronization. To learn this memory-augmented model effectively, we devise a three-stage training recipe: (1) adapting a bidirectional audio-video teacher to block-wise causal attention with simulated memory, (2) optimizing the memory construction and interaction pipeline through end-to-end distillation, and (3) applying online reinforcement post-training tailored for streaming audio-video generation. 
As a result, Ripple achieves $\sim$28 FPS at 480P resolution, over $15\times$ faster than the teacher, while capable of coherent long-form generation. Extensive experiments on both short-video and long-video benchmarks demonstrate our superior performance over existing offline and online joint audio-video generation methods.
\end{abstract}


\begin{figure}[t]\centering\includegraphics[width=0.48\textwidth]
{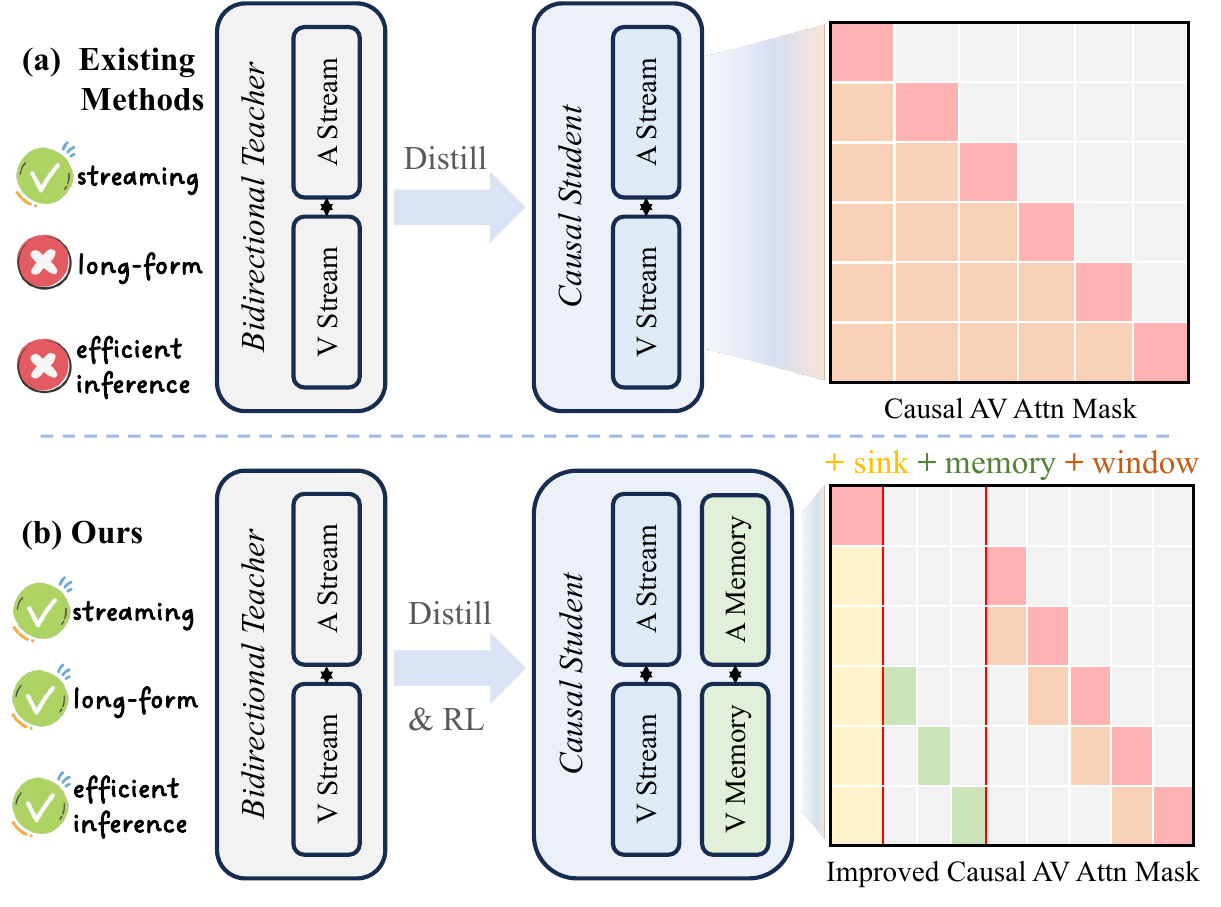}\vspace{-5pt}
\caption{\textbf{Motivation.} Existing streaming audio-video methods rely on an ever-growing KV cache, leading to increasing inference cost. Our Ripple combines a fixed-length sliding window with cross-modal recurrent memory for constant-cost inference and long-range coherence. Additionally, we introduce reinforcement post-training for joint generation, a key training difference from prior streaming AV approaches.}\label{fig:motivation}\end{figure}

\section{Introduction}
In the past few years, video generation has advanced rapidly. Models such as Wan~\cite{wan2025wan}, HunyuanVideo~\cite{wu2025hunyuanvideo}, and CogVideoX~\cite{yang2025cogvideox} achieve impressive visual quality and strong text alignment. However, these methods operate in an offline and bidirectional manner, requiring full temporal context before generating any output. This makes them unsuitable for real-time scenarios.

To overcome this limitation, recent work such as self-forcing~\cite{huang2026self-forcing} has explored distilling bidirectional diffusion models into causal counterparts capable of streaming generation with only a few denoising steps. While effective for video, these approaches leave the audio modality entirely unaddressed. More recently, streaming audio-video methods~\cite{su2026omniforcing,li2026hallo} fill this gap by extending self-forcing style causal distillation to a dual-stream architecture, as illustrated in Figure~\ref{fig:motivation}. During causal generation, these models attend to the entire history through an ever-growing KV cache. While effective for short sequences, the increasing attention context incurs growing computational cost and inference latency, making long-form streaming generation impractical. This raises a key question: \emph{how can we achieve real-time joint audio-video generation that is faster, more efficient, and capable of long-form generation?}

A natural solution is to adopt a fixed-length attention window to maintain constant-cost inference. However, truncating historical context inevitably discards long-range temporal and cross-modal information, making it difficult to preserve temporal coherence and audio-visual synchronization in a causal generation paradigm. Therefore, an effective streaming audio-video generator requires a compact memory mechanism that can continuously summarize historical multimodal context. Yet such a memory mechanism for streaming audio-video generation remains largely unexplored.

To this end, we propose \textbf{Ripple}, a real-time streaming audio-video generation system equipped with a cross-modal memory mechanism. Specifically, we introduce fixed-length modality-specific memory states that continuously summarize past audio and video context through memory attention and EMA-based~\cite{hunter1986ema} updates. To facilitate cross-modal synchronization, the two memory states further exchange information via cross-modal attention, producing enriched multimodal representations that serve as long-term memory. During generation, these memory representations are prepended to the fixed-length sliding-window KV cache. In this way, Ripple preserves long-range consistency and cross-modal synchronization without sacrificing efficiency.

To effectively train such a memory-augmented dual-stream model, we introduce a three-stage training paradigm. First, to bridge the inherent gap between offline full-context generation and online block-wise causal inference, we adapt the bidirectional attention pattern to causal window attention with simulated memory obtained from inference rollout. This allows the model to gradually learn the new causal attention pattern, even before the memory system is fully functional.  The second stage optimizes the complete memory construction and interaction pipeline through memory-forcing distillation. During this stage, the model learns to effectively leverage accumulated cross-modal context for temporally coherent generation. Finally, to further improve the overall generation performance, we apply online reinforcement post-training~\cite{xue2025dancegrpo} adapted for joint streaming audio-video generation, with reward signals targeting audio-visual quality,  synchronization, and speech alignment.

Our contributions are threefold. (1) We develop a three-stage training paradigm that effectively transfers a bidirectional audio-video generation teacher into a causal streaming generator, comprising three stages: memory-augmented block-causal adaptation, cross-modal memory-forcing distillation, and online dual-stream reinforcement post-training. (2) We introduce a cross-modal recurrent memory mechanism following a construct-then-interact paradigm, enabling constant-cost streaming inference while preserving long-range temporal consistency and cross-modal coherence. (3) We build a real-time streaming audio-video generation system achieving $\sim$28 FPS at 480P on a single NVIDIA H100 GPU, over 15$\times$ faster than the teacher. Experiments on VerseBench~\cite{wang2025universe} and our 30-second long-video benchmark demonstrate Ripple’s superior performance. 

\section{Related Work}
\paragraph{Offline Video Generation.}
Diffusion-based~\cite{song2020ddim} video generation models have achieved substantial progress in visual quality and text alignment. Most existing approaches adopt an offline, bidirectional paradigm, ranging from early models such as SVD~\cite{blattmann2023svd} and CogVideoX~\cite{yang2025cogvideox} to more recent systems including HunyuanVideo~\cite{wu2025hunyuanvideo}, Wan~\cite{wan2025wan}, and Sora~\cite{openai2024sora}. Building upon these advances in visual generation, unified multimodal generation has also progressed rapidly, with joint audio-video models such as LTX~\cite{hacohen2024ltx,hacohen2026ltx2}, UniVerse~\cite{wang2025universe}, JavisDiT~\cite{liu2025javisdit,liu2026javisdit++}, MOVA~\cite{team2026mova}, and Seedance~\cite{seedance2026seedance}. Despite their impressive generation quality, these models remain computationally expensive and suffer from high inference latency, motivating the development of real-time streaming joint audio-video generation.

\paragraph{Online Video Generation.}
To reduce latency, various distillation techniques have been proposed to compress diffusion sampling steps.  
Distribution Matching Distillation (DMD)~\cite{yin2024dmd} minimizes an approximate KL divergence to align the distribution between student and teacher, whereas Consistency Models~\cite{song2023consistency} enforce self-consistency along probability-flow ODE trajectories to learn a single-step mapping. 
Although these methods accelerate generation, they are built upon bidirectional architectures and are not suitable for streaming generation. 
CausVid~\cite{yin2025causvid} first introduced a streaming diffusion framework by distilling a bidirectional video teacher into a causal student through asymmetric DMD. Self-Forcing~\cite{huang2026self-forcing} further alleviated the exposure bias inherent in autoregressive video generation by training the model on its own rollouts to simulate error accumulation during inference. Causal-Forcing~\cite{zhu2026causalforcing} extended this direction with stricter causal constraints by first training a causal teacher. However, these methods are restricted to the video modality. Recently, OmniForcing~\cite{su2026omniforcing} and Hallo-Live~\cite{li2026hallo} enabled real-time streaming audio-video generation by adapting self-forcing to dual-stream, but they are limited to short horizon ($\sim$5 seconds). In contrast, Ripple supports long-form streaming generation through a cross-modal memory construction and interaction.

\section{Method}

\paragraph{Overview.}
We present Ripple, a streaming audio-video generation system that generates audio and video in a causal, block-by-block manner. 
To maintain long-term dependency under streaming constraints, Ripple introduces a cross-modal recurrent memory mechanism, combining modality-specific memory construction with cross-modal interaction. Based on this design, we further develop a three-stage training paradigm. Finally, we describe the streaming inference pipeline that enables real-time constant-cost generation.

\begin{figure*}[t]\centering\includegraphics[width=0.99\textwidth]{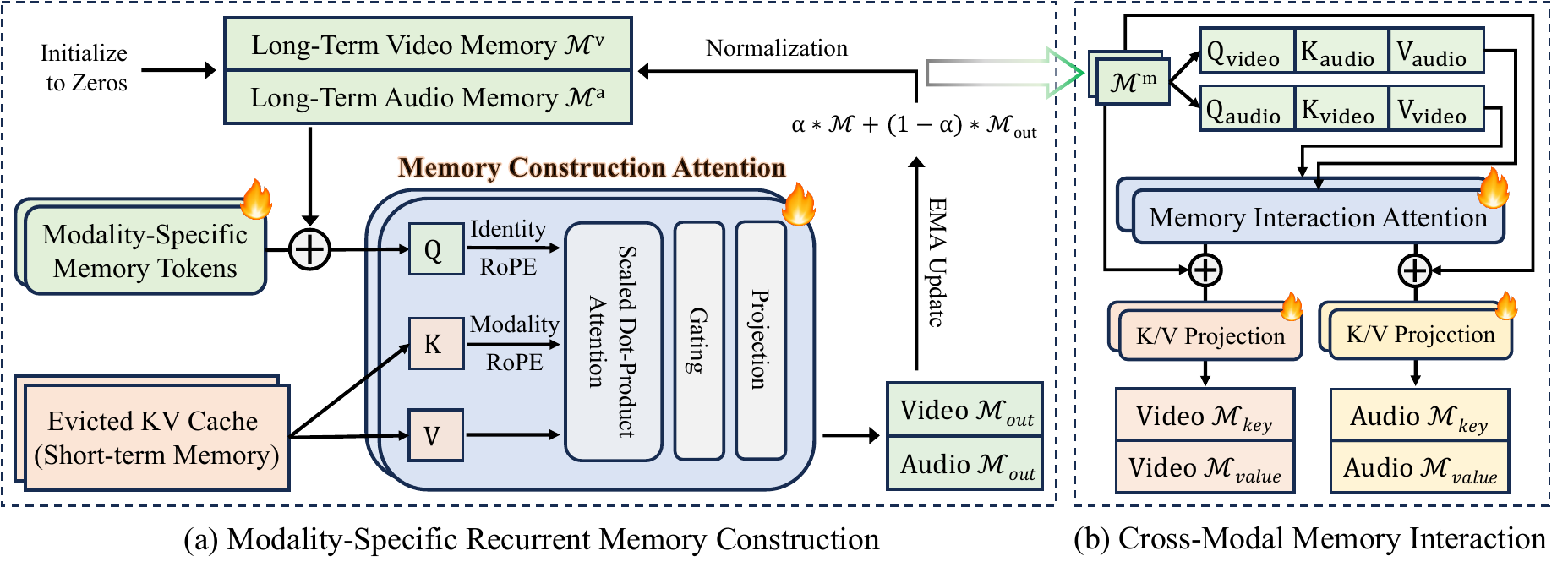}
\vspace{-10pt}
\caption{\textbf{Illustration of Cross-modal Recurrent Memory Mechanism.} (a) Modality-specific recurrent memory construction: audio and video memory states are independently updated to summarize historical context. (b) Cross-modal memory interaction: the two memory streams exchange information before being injected as key-value pairs into transformer layers.}
\label{fig:memory_mechanism}\end{figure*}

\subsection{Cross-Modal Recurrent Memory}
Long-form streaming generation requires not only a fixed-length context window for efficiency, but also preservation of long-term historical information. Therefore, we design a cross-modal recurrent memory mechanism that augments a sliding-window attention scheme. As shown in Figure~\ref{fig:memory_mechanism}, the attention context is composed of three parts in a sequential manner: (i) the first-block KV cache, which provides an initial global anchor, (ii) long-term cross-modal memory, which summarizes historical information beyond the window, and (iii) the previous-block KV cache, which captures the most recent context. Based on this formulation, we next describe the construction and interaction of our long-term memory.

\paragraph{Modality-Specific Memory Construction.}

During inference rollout, when a block's KV representations are evicted from the sliding context window, they are used to update the corresponding modality's memory state. For each modality $m \in \{v, a\}$, a set of $N_m$ learnable memory queries $\mathbf{Q}_m \in \mathbb{R}^{N_m \times d_m}$ attend over the evicted key-value pairs $(\mathbf{K}_{\text{evict}}^m, \mathbf{V}_{\text{evict}}^m)$ to produce a memory readout $\mathbf{M}_{\text{out}}^m$. Let
$\hat{\mathbf{Q}}_m = \mathcal{R}(\mathbf{Q}_m + \mathbf{M}^m, \mathbf{0})$ and $\hat{\mathbf{K}}_m = \mathcal{R}(\mathbf{K}_{\text{evict}}^m, \mathbf{p}^m)$, where $\mathcal{R}(\mathbf{x}, \mathbf{p})$ denotes the RoPE~\cite{su2024rope} operation at position $\mathbf{p}$, and $\mathbf{M}^m$ represents the modality-specific memory state.
The memory readout is formulated:
\begin{equation}
    \mathbf{M}_{\text{out}}^m = \text{softmax}\!\left(\frac{\hat{\mathbf{Q}}_m\, \hat{\mathbf{K}}_m^\top}{\sqrt{d^m}}\right)\mathbf{V}_{\text{evict}}^m,
\end{equation}
where $d^m$ is the attention dimension of modality $m$. $\mathbf{Q}_m$ are added to $\mathbf{M}^m$, enabling the readout to leverage both the learnable tokens and historical context. The memory queries are encoded with zero-position RoPE, making them position-agnostic. The evicted keys $\mathbf{K}_{\text{evict}}^m$ retain their original modality-specific RoPE encodings at positions $\mathbf{p}^m$, preserving the spatial-temporal structure of the source context. This asymmetric design allows the memory to freely aggregate information without being anchored to any specific location.

The memory state $\mathbf{M}^m$ is then updated by incorporating $\mathbf{M}_{\text{out}}^m$ via Exponential Moving Average (EMA):
\begin{equation}
    \mathbf{M}^m \leftarrow \text{norm}\!\left(\alpha\, \mathbf{M}^m + (1 - \alpha)\, \mathbf{M}_{\text{out}}^m\right),
    \label{eq:memory_update}
\end{equation}
where $\alpha \in (0, 1)$ is the ratio controlling the balance between past memory and new context, and $\text{norm}$ denotes L2 normalization applied to stabilize the memory magnitude during long-form inference. This recurrent update allows each of the $N_m$ memory states to self-evolve and accumulate compressed context over historical sequences.

\paragraph{Cross-Modal Memory Interaction.}
To further enhance cross-modal synchronization in the learned memory space, we introduce cross-modal attention between the two memory streams after the memory construction. Since video and audio memories reside in different feature spaces, we first project them into a shared dimension $d_c$ via learnable matrices $\mathbf{W}^Q, \mathbf{W}^K, \mathbf{W}^V$. Let $\mathbf{Q}^m = \mathbf{W}_m^Q\mathbf{M}^m$, $\mathbf{K}^m = \mathbf{W}_m^K\mathbf{M}^m$, $\mathbf{V}^m = \mathbf{W}_m^V\mathbf{M}^m$ for $m \in \{v, a\}$. The memory states then exchange information, following:
\begin{align}
    \tilde{\mathbf{M}}^v &= \mathbf{M}^v + \text{softmax}\!\left(\frac{\mathbf{Q}^v (\mathbf{K}^a)^\top}{\sqrt{d_c}}\right)\mathbf{V}^a, 
    \\
    \tilde{\mathbf{M}}^a &= \mathbf{M}^a + \text{softmax}\!\left(\frac{\mathbf{Q}^a (\mathbf{K}^v)^\top}{\sqrt{d_c}}\right)\mathbf{V}^v.
    \label{eq:memory_interaction}
\end{align}
The residual formulation preserves the modality-specific content while allowing each memory stream to incorporate complementary information from the other modality. The interacted representations $\tilde{\mathbf{M}}^v$ and $\tilde{\mathbf{M}}^a$ are then passed through modality-specific linear projection layers to produce the final long-term memory key-value pairs $(\mathbf{K}_{\text{mem}}^m, \mathbf{V}_{\text{mem}}^m)$, which are used in self-attention. This enables each block to attend to a compact summary of long-range historical context from both modalities, without increasing the per-step sequence length.

\begin{figure*}[t]
\centering
\includegraphics[width=0.99\textwidth]{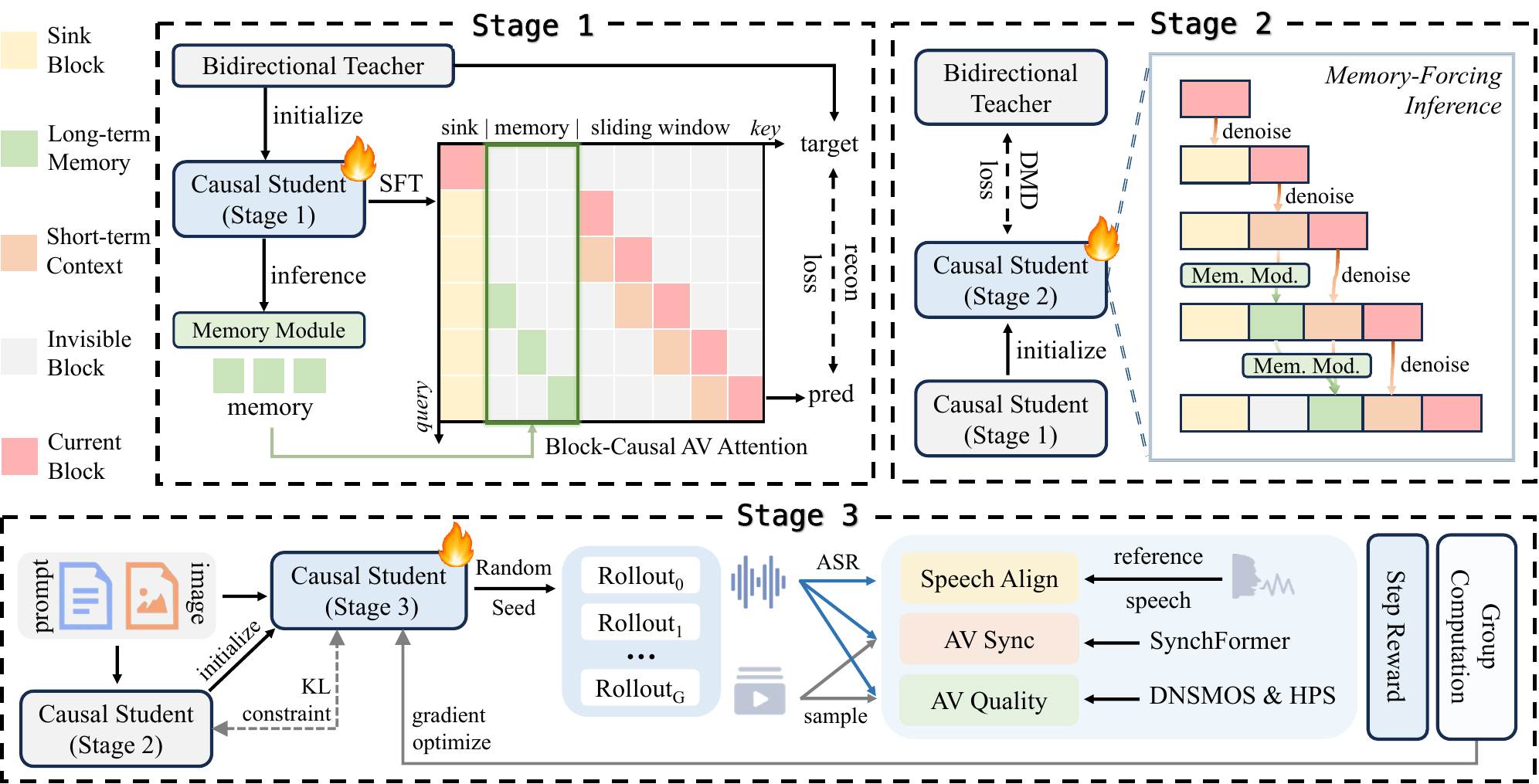}\vspace{-3pt}
\caption{\textbf{Illustration of our Three-Stage Training Recipe.} (1) Memory-augmented block-causal mask adaptation: the bidirectional teacher backbone is adapted to causal, block-by-block attention with simulated memory derived from inference rollout, allowing the model to learn the new causal attention pattern. (2) Cross-modal memory-forcing distillation: the complete model is trained end-to-end under a distillation objective, to leverage accumulated cross-modal context for temporally coherent generation. (3) Online diffusion-based reinforcement: the streaming model is post-trained with a composite reward of audio-visual quality, synchronization, and speech alignment, further improving perceptual performance and modality mutual alignment.}
\label{fig:training_stages}
\end{figure*}

\subsection{Progressive Training Stages}

To effectively exploit our long-term memory, we design a three-stage training recipe consisting of: (1) memory-augmented block-causal adaptation to learn the new attention pattern, (2) memory-forcing distillation to approximate teacher performance under streaming constraints, and (3) online dual-stream diffusion reinforcement to further optimize towards perceptual quality and cross-modal alignment.

\paragraph{Memory-Augmented Block-Causal Adaptation.}

As illustrated in Figure~\ref{fig:training_stages}(a), the first stage bridges the gap between the full-context bidirectional model and the local-context streaming setting. Specifically, we replace the bidirectional attention with a memory-augmented block-causal attention under sliding window, where each block attends to the sink tokens (i.e., the first-block KV cache), the long-term cross-modal memory, and the immediately preceding block KV cache.
We simulate the long-term cross-modal memory via inference: the model first performs causal rollout to obtain intermediate KV representations, which are then reused as placeholder memory inputs. The produced memory representations are detached from the computational graph and are used only to emulate memory usage. This exposes the model to the memory-augmented causal attention pattern and allows it to learn how to utilize memory representations. Training is conducted under the diffusion-forcing~\cite{chen2024diffusion-forcing} framework with a flow matching objective~\cite{lipman2022flow} against the teacher's ODE trajectories. This provides a favorable initialization for the subsequent distillation stage. The training data preparation is detailed in the Supplement.

\paragraph{Memory-Forcing Training.}

Building upon the adapted backbone, we then optimize the complete construct-then-interact memory pipeline end-to-end, as shown in Figure~\ref{fig:training_stages}(b). We distill a four-step student model from the bidirectional teacher using the DMD~\cite{yin2024dmd} objective. Each generation block of the student model is conditioned on the first clean block, cross-modal memory key-value representations, and the latest generated clean block. During training, memory states are continuously updated (Equation~\ref{eq:memory_update}), with cross-modal information exchanged between modalities to produce cross-modal memory key-value representations (Equation~\ref{eq:memory_interaction}). This memory-forcing mechanism narrows the gap between causal streaming inference and bidirectional generation while maintaining a constant-cost inference speed and long-horizon visual and timbre consistency.

\paragraph{Online Diffusion-based Reinforcement.}

Although the distillation successfully transfers generation capability from the teacher to the streaming model, it does not directly optimize towards perceptual quality. To further improve speech alignment, audio-visual synchronization, and overall generation quality, we introduce an online dual-stream diffusion reinforcement post-training stage. 
To the best of our knowledge, we are the first to apply such a reinforcement learning strategy to streaming audio-video generation.

Thanks to the streaming inference, online rollouts can be performed efficiently. We adopt a GRPO-based framework~\cite{xue2025dancegrpo}, where each GPU rank generates one dual-stream self-forcing rollout with a different seed, and rewards are aggregated across ranks to compute group-relative advantages.
For each rollout, we define a composite reward:
\begin{equation}
r =
w_1 r_{\mathrm{av}}
+
w_2 r_{\mathrm{sp}}
+
w_3 r_{\mathrm{vid}}
+
w_4 r_{\mathrm{aud}},
\end{equation}
where $r_{\mathrm{av}}$ measures audio-visual synchronization using Synchformer~\cite{iashin2024synchformer}, $r_{\mathrm{sp}}$ measures speech alignment using ASR-based~\cite{radford2023robust} word and character error rates computed between the transcribed speech and the reference speech, $r_{\mathrm{vid}}$ evaluates video quality using HPS~\cite{wu2023hps} , and $r_{\mathrm{aud}}$ assesses audio quality via DNSMOS~\cite{reddy2021dnsmos}.

The rewards collected from all parallel rollouts are normalized to obtain group-relative advantages: $A_i=\frac{r_i-\mu_r}{\sigma_r}$,
where $\mu_r$ and $\sigma_r$ denote the mean and standard deviation of rewards within the rollout group.
The policy is then optimized by maximizing advantage-weighted diffusion transition probabilities with KL regularization, following:

\begin{align}
\mathcal{L}_{\mathrm{RL}}
=&
-
\mathbb{E}_{\tau_i}
\left[
A_i
\sum_t
\log p_{\theta}(x_{t-1}\mid x_t)
\right]
\nonumber\\
&+
\beta
D_{\mathrm{KL}}
\left(
\pi_{\theta}
\Vert
\pi_{\mathrm{ref}}
\right),
\end{align}

where $\log p_{\theta}(x_{t-1}\mid x_t)$ denotes the diffusion transition log-probability, $\pi_{\theta}$ and $\pi_{\mathrm{ref}}$ denote the current policy and the frozen reference policy, respectively.  $\beta$ controls the strength of KL regularization. The KL term stabilizes optimization by constraining deviation from the pretrained policy.
This reinforcement stage operates on the joint audio-video generation policy. Specifically, the log-probability $\log p_{\theta}(x_{t-1}\mid x_t)$ is computed as the average of the audio and video diffusion transition log-probabilities at each denoising step (i.e., 0.5 for each modality).
To balance alignment and perceptual quality, we employ a progressive reward schedule. At the beginning, reinforcement learning focuses on speech alignment and audio-visual synchronization. Next, the optimization target gradually shifts toward perceptual audio and video quality while retaining a small alignment reward. This progressive reinforcement strategy can jointly improve speech alignment, audio-visual synchronization, and generation fidelity.

\subsection{Streaming Inference}

During inference, Ripple generates audio and video in a causal, block-by-block manner. This design enables incremental streaming output, where each block can be immediately emitted upon generation. It also enables efficient generation under a bounded context, achieving constant-cost per-block generation while preserving long-range cross-modal information. To generate durations beyond the training horizon, we maintain a fixed-size KV cache and adopt a rolling cache strategy, following Self-Forcing++~\cite{cui2025selfforcing++}. In addition, RoPE positions are clipped to the maximum training position whenever they exceed it. For interactive generation, when switching prompts, we reset both audio and video memory states to zero to prevent interference from previous contexts. In this way, Ripple is capable of generating minute-level audio and video with long-form consistency.

\section{Experiment}

\begin{table*}[!t]
\centering
\resizebox{\textwidth}{!}{
\begin{tabular}{lcccccccc}
\toprule
\multirow{2}{*}{\textbf{Model}} & \multirow{2}{*}{\textbf{Latency\(\downarrow\)}} & \multicolumn{3}{c}{\textbf{Audio-Speech}} & \multicolumn{2}{c}{\textbf{Audio-Video}} & \multicolumn{2}{c}{\textbf{Lip-Sync}} \\
\cmidrule(lr){3-5} \cmidrule(lr){6-7} \cmidrule(lr){8-9}
 & & IS\(\uparrow\) & CLAP\(\uparrow\) & DNSMOS\(\uparrow\) & AV-Align\(\downarrow\) & IB-Score\(\uparrow\) & LSE-D\(\downarrow\) & LSE-C\(\uparrow\) \\
\midrule
Ovi-1.1~\cite{low2025ovi} & 112s & 1.062 & 0.221 & 3.480 & \textbf{0.193} & 0.231 & 8.816 & 5.296 \\
MOVA-360P~\cite{team2026mova} & 656s & \underline{1.112} & 0.208 & \underline{3.838} & 0.226 & 0.344 & 8.126 & 5.979 \\
UniVerse-1~\cite{wang2025universe} & 143s & 1.039 & 0.182 & 3.485 & 0.232 & 0.255 & 11.347 & 2.433 \\
DaVinci-Base~\cite{chern2026davinci} & 43s & \textbf{1.117} & 0.195 & 3.579 & \underline{0.218} & 0.321 & 7.794 & 5.466 \\
LTX-2.3 (Teacher) ~\cite{hacohen2026ltx2} & 74s & 1.103 & \textbf{0.236} & 3.742 & 0.223 & \underline{0.397} & \textbf{6.965} & \textbf{6.371} \\
\rowcolor{gray!20}
\textbf{Ripple (Ours)} & \textbf{5.9s} & 1.085 & \underline{0.225} & \textbf{3.858} & 0.219 & \textbf{0.410} & \underline{7.792} & \underline{6.078} \\
\midrule
OmniForcing~\cite{su2026omniforcing} & \underline{7.7s} & \textbf{1.210} & \underline{0.147} & \underline{3.590} & \underline{0.275} & \underline{0.181} & \underline{12.868} & \underline{1.419} \\
\rowcolor{gray!20}
\textbf{Ripple (Ours)} & \textbf{4.2s} & \underline{1.036} & \textbf{0.209} & \textbf{3.809} & \textbf{0.260} & \textbf{0.393} & \textbf{8.701} & \textbf{5.727} \\
\midrule
Hallo-Live~\cite{li2026hallo} & \underline{5.4s} & \underline{1.021} & \textbf{0.224} & \underline{3.573} & \textbf{0.265} & \underline{0.186} & \underline{10.241} & \underline{4.277} \\
\rowcolor{gray!20}
\textbf{Ripple (Ours)} & \textbf{3.9s} & \textbf{1.084} & \underline{0.221} & \textbf{3.764} & \underline{0.274} & \textbf{0.369} & \textbf{8.793} & \textbf{5.480} \\
\bottomrule
\end{tabular}
}\vspace{-3pt}
\caption{\textbf{Evaluation on Short Videos.} Metrics are reported on the VerseBench~\cite{wang2025universe} Set3 subset. Our Ripple is compared against (1) offline audio-video generation methods: it achieves comparable performance against the teacher, with SOTA results on several metrics (e.g., DNSMOS, IB-Score); (2) online streaming methods: Ripple outperforms OmniForcing and Hallo-Live across most metrics with lower latency. \textbf{Bold} indicates the best, \underline{underline} indicates the second best.}
\label{tab:verse_bench_evalution}
\end{table*}

\begin{table*}[htbp]
\centering
\small
\resizebox{\textwidth}{!}{
\begin{tabular}{lccccc}
\toprule
\textbf{Model} & 
\textbf{Latency$\downarrow$} &
\textbf{AV Quality$\uparrow$} & 
\textbf{Speech Align$\uparrow$} & 
\textbf{ID Consistency$\uparrow$} & 
\textbf{AV Synchronize$\uparrow$} \\
\midrule
Teacher: LTX-2.3~\cite{hacohen2026ltx2}  & 332s & 0.736 & \textbf{0.464} & 0.914 & \textbf{0.608} \\
\rowcolor{gray!20}
Student: Ripple (Ours)  & \textbf{23s} & \textbf{0.751} & 0.405 & \textbf{0.944} & 0.579 \\
\bottomrule
\end{tabular}
}\vspace{-3pt}
\caption{\textbf{Evaluation on Long Videos.} Although trained only on 7-second videos, our Ripple outperforms its teacher LTX-2.3 in audio-visual quality and identity consistency on the 30-second generation, while achieving about 15$\times$ speedup in latency.}
\label{tab:long_video_evalution}
\end{table*}

\begin{table*}[ht!]
    \centering
    \begin{minipage}{0.490\textwidth}
    \resizebox{\textwidth}{!}{
    \begin{tabular}{lccccc}
        \toprule
        \textbf{Training Stage} & \multicolumn{5}{c}{\textbf{Choice}} \\
        \midrule
        Stage 1 & {\ding{52}} & {\ding{52}} & {\ding{52}} & {\ding{52}} & \textcolor[RGB]{192, 192, 192}{\ding{56}} \\
        Stage 2 & {\ding{52}} & \textcolor[RGB]{192, 192, 192}{\ding{56}} & {\ding{52}} & \textcolor[RGB]{192, 192, 192}{\ding{56}} & {\ding{52}} \\
        Stage 3 & {\ding{52}} & \textcolor[RGB]{192, 192, 192}{\ding{56}} & \textcolor[RGB]{192, 192, 192}{\ding{56}} & {\ding{52}} & {\ding{52}} \\
        \midrule
        {AV Quality$\uparrow$} & \textbf{0.751} & 0.641 & \underline{0.738} & 0.651 & 0.576 \\
        {Speech Align$\uparrow$} & \textbf{0.405} & 0.358 & \underline{0.386} & 0.364 & 0.257 \\
        {ID Consistency$\uparrow$} & \textbf{0.944} & 0.905 & \underline{0.939} & 0.907 & 0.887 \\
        {AV Synchronize$\uparrow$} & \textbf{0.579} & 0.545 & \underline{0.553} & 0.568 & 0.528 \\
        \bottomrule
    \end{tabular}
    }\vspace{-3pt}
    \caption{\textbf{Ablation of Training Recipe.} Our full three-stage training recipe consistently achieves the best performance across all metrics on our long-video benchmark. Each stage is beneficial, with the first stage being the most important.}
    \label{table: ablation_training}
    \end{minipage}%
    \hfill
    \begin{minipage}{0.472\textwidth}
    \centering
    \resizebox{\textwidth}{!}{
    \begin{tabular}{lcccc}
        \toprule
        \textbf{Component} & \multicolumn{4}{c}{\textbf{Choice}} \\
        \midrule
        Memory Construction & {\ding{52}} & {\ding{52}} & {\ding{52}} & \textcolor[RGB]{192, 192, 192}{\ding{56}} \\
        Memory Interaction & {\ding{52}} & \textcolor[RGB]{192, 192, 192}{\ding{56}} & {\ding{52}} & \textcolor[RGB]{192, 192, 192}{\ding{56}} \\
        Memory Normalization & {\ding{52}} & {\ding{52}} & \textcolor[RGB]{192, 192, 192}{\ding{56}} & \textcolor[RGB]{192, 192, 192}{\ding{56}} \\
        \midrule
        {AV Quality$\uparrow$} & \textbf{0.751} & \underline{0.734} & 0.592 & 0.684 \\
        {Speech Align$\uparrow$} & \textbf{0.405} & \underline{0.399} & 0.287 & 0.381 \\
        {ID Consistency$\uparrow$} & \underline{0.944} & \textbf{0.947} & 0.875 & 0.910 \\
        {AV Synchronize$\uparrow$} & \textbf{0.579} & 0.563 & 0.532 & \underline{0.565} \\
        \bottomrule
    \end{tabular}
    }\vspace{-3pt}
    \caption{\textbf{Ablation of Memory Mechanism.} Memory construction and interaction improve overall performance. Normalization is critical for training stability. Removing memory substantially degrades long-video generation performance.}
    \label{table: ablation_memory}
    \end{minipage}
\end{table*}

\begin{table*}[!t]
\centering
\small
\resizebox{\textwidth}{!}{
\begin{tabular}{p{1.6cm}p{0.7cm}p{0.7cm}p{0.7cm}cccccc}
\toprule
\textbf{Setting} & \textbf{$N_{v}$} & \textbf{$N_{a}$} & \textbf{$\alpha$} & \textbf{Latency$\downarrow$} & 
\textbf{AV Quality$\uparrow$} & 
\textbf{Speech Align$\uparrow$} & 
\textbf{ID Consistency$\uparrow$} & 
\textbf{AV Synchronize$\uparrow$} \\
\midrule
$\#1$ & 256 & 16 & 0.9 & \textbf{23.03s} & 0.735 & 0.383 & 0.937 & 0.560 \\
$\#2$ & 256 & 32 & 0.9 & \underline{23.12s} & \underline{0.747} & \underline{0.401} & 0.942 & \underline{0.574} \\
\rowcolor{gray!20}
\textbf{$\#3$ (Ours)} & \textbf{512} & \textbf{32} & \textbf{0.9} & 23.35s & \textbf{0.751} & \textbf{0.405} & \textbf{0.944} & \textbf{0.579} \\
$\#4$ & 512 & 32 & 0.5 & 23.35s & 0.746 & 0.392 & \underline{0.943} & 0.569 \\
$\#5$ & 512 & 32 & 0.1 & 23.35s & 0.738 & 0.381 & 0.932 & 0.561 \\
$\#6$ & 512 & 32 & 1.0 & 23.35s & 0.734 & 0.389 & 0.930 & 0.562 \\
\bottomrule
\end{tabular}
}\vspace{-3pt}
\caption{\textbf{Ablation of Key Hyperparameters.} We study the effects of video memory tokens ($N_{v}$), audio memory tokens ($N_{a}$), and the EMA ratio ($\alpha$). All three hyperparameters affect long-video generation performance. Among the evaluated configurations, $N_{v}=512$, $N_{a}=32$, and $\alpha=0.9$ achieve the best overall performance under streaming inference constraint.}
\label{tab:ablation_hyperparameter}
\end{table*}

\paragraph{Training Details.} Our teacher is LTX-2.3~\cite{hacohen2026ltx2}, a 19B open-sourced joint audio-video generative model. The training duration is 7s, and the resolution is dynamically sampled with an average aspect ratio corresponding to approximately $384 \times 672$. The EMA ratio $\alpha$ of memory is set to 0.9, with $N_v = 512$ video memory query tokens and $N_a = 32$ audio memory query tokens. All experiments are conducted on 32 NVIDIA H100 GPUs. Additional implementation details are provided in the Supplement.

\paragraph{Evaluation Details.}
We adopt the VerseBench~\cite{wang2025universe} Set-3 subset, which contains 100 human-centered cases and serves as a standard benchmark for state-of-the-art evaluation. For all methods, we report results under their default configurations, with the seed fixed to 42 for reproducibility. Since this benchmark is limited to short videos, we further construct a long-video benchmark consisting of 50 prompts (25 in Chinese and 25 in English). More details about the long-video benchmark are in the Supplement.

\subsection{Comparison Results}

We compare Ripple with both offline and online audio-video generation methods under quantitative and qualitative evaluations. The results show that Ripple demonstrates strong performance while achieving real-time inference efficiency. We conduct an additional user study in the Supplement.

\paragraph{Quantitative Results on Short Videos.}
As shown in Table~\ref{tab:verse_bench_evalution},
compared with offline methods, Ripple achieves over $15\times$ lower latency while maintaining comparable generation quality. Specifically, Ripple achieves the best DNSMOS and IB-Score, and ranks second on multiple remaining metrics. These results demonstrate that Ripple preserves the generation quality of the bidirectional teacher while substantially improving inference efficiency. For online methods, we compare against OmniForcing and Hallo-Live using their first frames and under the same setting for fair comparison. Ripple achieves both higher inference efficiency and better overall generation quality. We attribute these gains to our cross-modal long-term memory, which enables effective context modeling under a fixed-length attention window.

\paragraph{Quantitative Results on Long Videos.} On our long-video benchmark, we generate 30-second videos and compare Ripple with its teacher. We attempted to evaluate streaming methods including OmniForcing and Hallo-Live, but they failed to generate videos of this duration. As shown in Table~\ref{tab:long_video_evalution}, Ripple achieves significantly faster inference (23s vs. 332s). Despite being trained only on 7-second videos, Ripple achieves better identity consistency and audio-visual quality, demonstrating the effectiveness of the proposed memory mechanism for long-horizon generation. We observe slight drops in speech alignment and audio-visual synchronization, which are acceptable considering the substantial speedup.

\begin{figure*}[!t]
\centering
\includegraphics[width=0.99\textwidth]{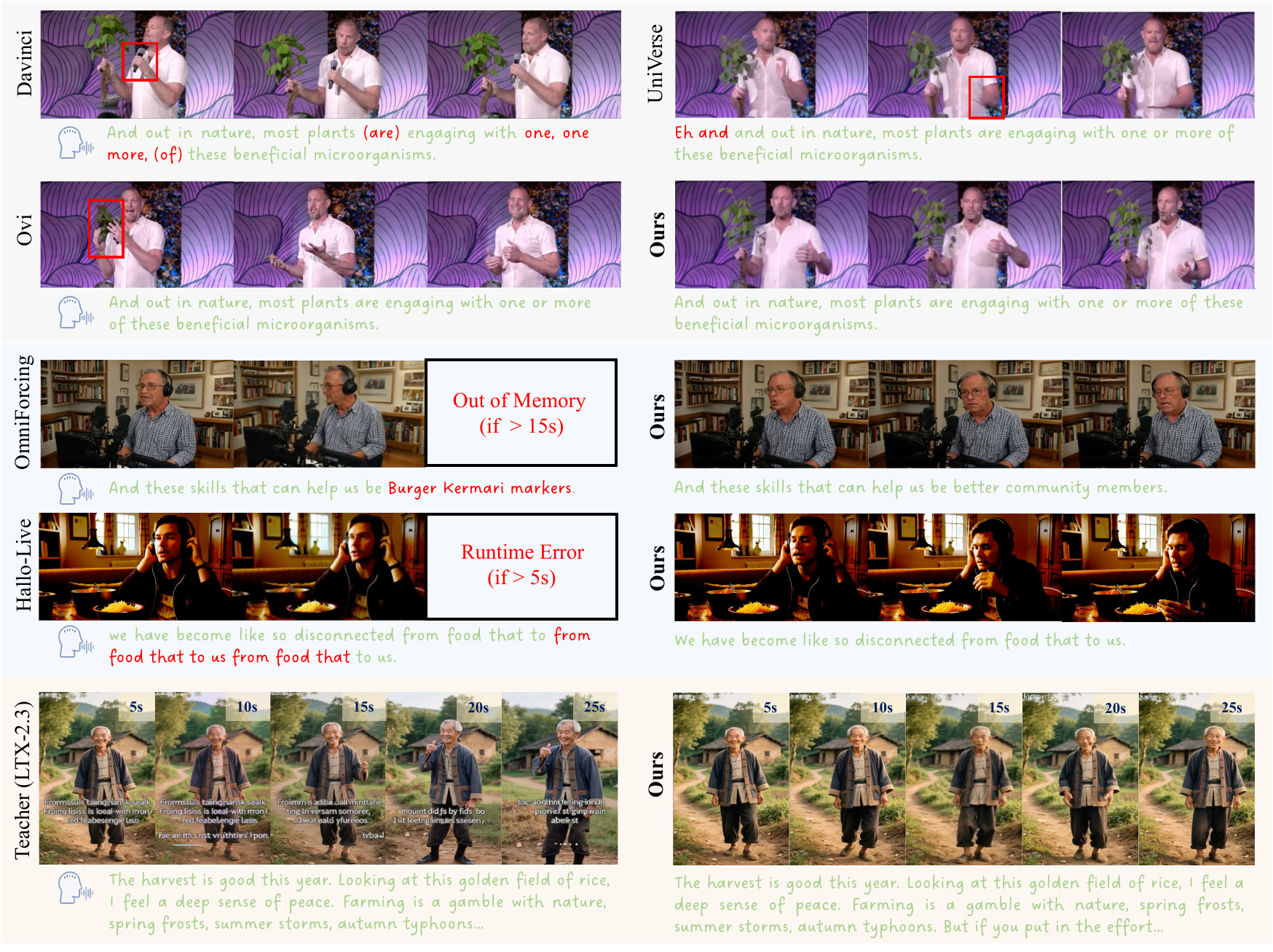}\vspace{-5pt}
\caption{\textbf{Qualitative Comparison.} Compared with existing offline and online streaming audio-video generation methods, Ripple achieves high-quality generation on short videos while enabling real-time streaming inference. On long videos, Ripple achieves better temporal consistency than the teacher, demonstrating the effectiveness of our memory mechanism. In the transcribed speech, \textcolor{red}{red} indicates mispronounced/incorrect words, \textcolor{green!50}{green} indicates correct ones, and (parentheses) denote omissions.}
\label{fig:compare}
\end{figure*}

\paragraph{Qualitative Comparison.}
As shown in Figure~\ref{fig:compare}, we compare Ripple with both offline and online audio-video generation methods. On short videos, Ripple achieves competitive performance compared with offline models while enabling real-time inference. Moreover, Ripple consistently outperforms existing streaming methods in speech accuracy and long-form generation. On long videos, Ripple demonstrates better temporal consistency than the teacher model, highlighting the effectiveness of our memory mechanism. Additional visualization results are in the Supplement.

\subsection{Ablation Studies} 

We conduct ablations on our long-video benchmark to validate the proposed three-stage training strategy, cross-modal recurrent memory mechanism, and key hyperparameter choices of memory. We provide an additional ablation of the reinforcement learning stage in the Supplement. 

\paragraph{Ablation of Training Stages.}

As shown in Table~\ref{table: ablation_training}, the complete three-stage recipe achieves the best performance across all metrics. Removing any stage leads to noticeable degradation, confirming that all three stages are essential. Among them, the first stage contributes the most, particularly in AV Quality and Speech Align, providing an important foundation for the subsequent training stages.

\paragraph{Ablation of Memory Mechanism.}
As shown in Table~\ref{table: ablation_memory}, memory construction contributes the most to generation quality and temporal consistency, leading to clear improvements in AV Quality and ID Consistency. Cross-modal memory interaction further boosts audio-visual quality, synchronization, and speech alignment. Memory normalization serves as the foundation for stable training. Removing it causes a dramatic performance collapse. Completely removing the entire memory yields bad performance, confirming its necessity. The full model achieves the best performance.

\paragraph{Ablation of Key Hyperparameters.}

Table~\ref{tab:ablation_hyperparameter} studies the effects of the memory query token numbers and the EMA memory ratio $\alpha$. Increasing the number of memory tokens slightly improves long-range consistency with minor impact on latency. We find that 512 video memory tokens and 32 audio memory tokens are sufficient. Varying $\alpha$ also affects generation quality, indicating that recurrent memory updates are functional. When $\alpha=1.0$, the memory is no longer updated, resulting in stale memory and degraded performance. When $\alpha=0.1$, the memory is updated too aggressively, making the memory representation unstable and noisy. In our experiments, $\alpha=0.9$ achieves the best performance.

\section{Conclusion}

We present Ripple, a streaming generation system that jointly synthesizes audio and video in a causal, block-by-block manner. At its core, Ripple introduces a cross-modal recurrent memory mechanism to maintain long-range temporal coherence and cross-modal synchronization. With sliding-window context, the generation computation maintains constant cost. Trained with a three-stage paradigm covering causal mask adaptation, memory-forcing distillation, and online dual-stream post-training, Ripple achieves comparable performance to the teacher while running at $\sim$28 FPS at 480P resolution, over 15$\times$ faster. We hope Ripple provides valuable insights for advancing real-time multi-modal generation.

\newpage
\appendix

\section{Preliminaries}

\subsection{Diffusion Transformer}

Diffusion Transformers (DiTs)~\cite{peebles2023scalable} provide a scalable alternative to convolutional diffusion backbones by formulating denoising as a sequence modeling problem. Following the latent diffusion paradigm, a pretrained VAE encoder first maps the input sample $x$ into a compact latent representation $z=E(x)$. The latent variable is subsequently perturbed with timestep-dependent Gaussian noise:
$z_t=\sqrt{\alpha_t}z+\sqrt{1-\alpha_t}\epsilon$,
where $\epsilon\sim\mathcal{N}(0,I)$ denotes the sampled noise and $\alpha_t$ controls the noise level.

The objective of DiT is to learn a parameterized denoising function that predicts the noise residual from the corrupted latent. Given a conditioning input $c$, the network $\epsilon_\theta(z_t,t,c)$ is optimized with the following reconstruction objective:
\begin{equation}
\mathcal{L}
=
\mathbb{E}_{t,z_t,c,\epsilon}
\left[
\left\|
\epsilon_\theta(z_t,t,c)-\epsilon
\right\|_2^2
\right].
\end{equation}

The transformer architecture enables DiT to perform large-scale generative modeling by replacing spatial convolutions with attention-based token interactions. Moreover, RoPE~\citep{su2024rope} provides position-dependent modulation without requiring fixed positional embeddings, allowing the model to better adapt to varying input configurations compared with U-Net~\cite{ronneberger2015unet}. RoPE applies the following rotation to each dimension pair $i$:
\begin{equation}
\label{rope}
R_i(x,m)=
\begin{bmatrix}
\cos(m\theta_i)&-\sin(m\theta_i)\\
\sin(m\theta_i)&\cos(m\theta_i)
\end{bmatrix}
\begin{bmatrix}
x_{2i}\\
x_{2i+1}
\end{bmatrix},
\end{equation}
where $m$ is the token position and $x$ represents the query or key feature. The rotation frequency is defined as $\theta_i=10000^{-2i/D}$, with $D$ being the attention hidden dimension. 

We adopt LTX-2.3-19B~\cite{hacohen2026ltx2} as the backbone diffusion transformer. It uses a dual-stream architecture to model audio and video, with separate decoders. Text conditions are encoded by a pretrained Gemma3~\cite{gemmateam2025gemma3technicalreport} encoder and injected via cross-attention.

\subsection{Distribution Matching Distillation}

Although diffusion models achieve impressive generation quality, their iterative denoising procedure introduces considerable inference latency. Distribution Matching Distillation (DMD)~\cite{yin2024dmd} addresses this issue by transferring the sampling behavior of a diffusion teacher into a generator capable of producing samples with a few denoising steps.

Specifically, let $G_\theta$ denote the student generator, which maps random noise $z\sim\mathcal{N}(0,I)$ into a generated sample $\hat{x}=G_\theta(z)$. Instead of directly matching individual samples, DMD encourages the student to reproduce the evolving data distribution of the teacher along the diffusion trajectory. At timestep $t$, the corresponding noisy distributions are represented as $p_{\mathrm{data},t}(x_t)$ and $p_{\theta,t}(x_t)$ for the teacher and student, respectively. The distribution alignment objective is formulated by minimizing the reverse KL divergence:
\begin{equation}
\mathcal{L}_{\mathrm{DMD}}
=
\mathbb{E}_t
\left[
D_{\mathrm{KL}}
(
p_{\theta,t}
\|
p_{\mathrm{data},t}
)
\right].
\end{equation}

In practice, DMD optimizes this objective through score discrepancy. The gradient update is determined by the difference between the real score and the fake score:
\begin{align}
\begin{aligned}
\nabla_\theta \mathcal{L}_{\mathrm{DMD}}
=
-\mathbb{E}_{t,z}
\Big[
&
(s_{\mathrm{real}}(x_t,t)
-
s_{\mathrm{fake},\phi}(x_t,t))^T \\
&\frac{\partial G_\theta(z)}{\partial\theta}
\Big].
\end{aligned}
\end{align}

Here, $s_{\mathrm{real}}$ represents the score function provided by the diffusion teacher, while $s_{\mathrm{fake},\phi}$ denotes the score estimator trained for the generated distribution. The noisy state $x_t$ is obtained by perturbing the generated sample through the forward diffusion process. The resulting student generator preserves the teacher distribution while requiring only a few sampling steps, enabling efficient inference. 

In our work, we build upon the standard DMD framework and further incorporate a cross-modal recurrent memory module, which enables long-form audio-video joint generation under real-time streaming constraints.

\section{Additional Training Details}

\subsection{Training Configuration}

\paragraph{Stage 1.}
We train the model using the AdamW optimizer~\citep{loshchilov2017decoupled} with $\beta_1=0.9$ and $\beta_2=0.999$. The learning rate is set to $2\times10^{-5}$ and the warmup steps are set to 100. Gradient accumulation is performed every 2 steps. Weight decay is set to 0.001. The total batch size is 64. The total training steps are 7,000. 

\paragraph{Stage 2.}
The generator (i.e., student) is optimized with AdamW using a learning rate of $2\times10^{-6}$, while the memory module uses a learning rate of $2\times10^{-5}$. The critic is optimized with a learning rate of $1\times10^{-7}$. All components use $\beta_1=0.9$ and $\beta_2=0.999$. The warmup steps are 100. Gradient accumulation is 1. The critic is updated 5 times per generator update. The total batch size is 32. The gradient clipping norm is set to 1.0. The total training steps are 5,000.

\paragraph{Stage 3.}
We adopt a two-phase reinforcement learning strategy. In the first phase, the model is trained for 200 steps using speech alignment and audio-visual synchronization rewards to improve cross-modal alignment, following:
\begin{equation}
\mathcal{R_\text{1}} = 0.5 \cdot r_{\mathrm{av}} + 0.5 \cdot r_{\mathrm{sp}}.
\label{eq:stage3_reward1}
\end{equation}
Then, the model is trained for another 200 steps to jointly optimize audio-visual quality with a composite reward:
\begin{equation}
\mathcal{R_\text{2}} = 0.1 \cdot (r_{\mathrm{av}} + r_{\mathrm{sp}}) + 0.2 \cdot r_{\mathrm{aud}} + 0.6 \cdot r_{\mathrm{vid}}.
\label{eq:stage3_reward2}
\end{equation}
We use the AdamW optimizer with a learning rate of $1\times10^{-6}$, $\beta_1=0.9$, and $\beta_2=0.999$. The warmup steps are set to 100. Gradient accumulation is set to 1. The group size is 32, and rewards are clipped to $[0, 1]$, advantages are clipped to the range $[-5, 5]$. A KL penalty coefficient of 0.1 is applied.

\subsection{Dataset Preparation}

Our dataset is collected from three sources. The first source consists of publicly available talking-head datasets, including HDTF~\cite{zhang2021flow}, VFHQ~\cite{xie2022vfhq}, VoxCeleb2~\cite{chung2018voxceleb2}, CelebV-Text~\cite{yu2023celebv}, and AVSpeech~\cite{ephrat2018looking}. The second source is OpenHumanVid~\cite{li2024openhumanvid}, which provides diverse human-centric videos collected from movies and TV shows. The third source is our proprietary high-quality Chinese and English talking-head dataset.

We apply quality filtering based on aesthetic scores, motion quality, blur detection, and audio quality to remove low-quality samples. After filtering, we obtain a large-scale audio-video dataset containing about 3M high-quality clips.

For Stage 1 training, we first use the teacher model to generate ODE trajectories. For each sample, we randomly select four timesteps along the ODE trajectory: 1000, 757, 522, and 0. Here, timestep 0 corresponds to the clean data, while the remaining timesteps represent intermediate noisy states. Each training sample consists of a prompt and a corresponding first-frame image as conditional inputs. Due to the computational cost of teacher's ODE trajectory extraction, Stage 1 uses a subset of approximately 4K sequences. For Stage 2 and Stage 3 training, we use the full dataset, where each training sample is a 6.5-second audio-video clip.

\section{Long-Video Benchmark}

\begin{figure*}[!t]
\centering
\includegraphics[width=0.99\textwidth]{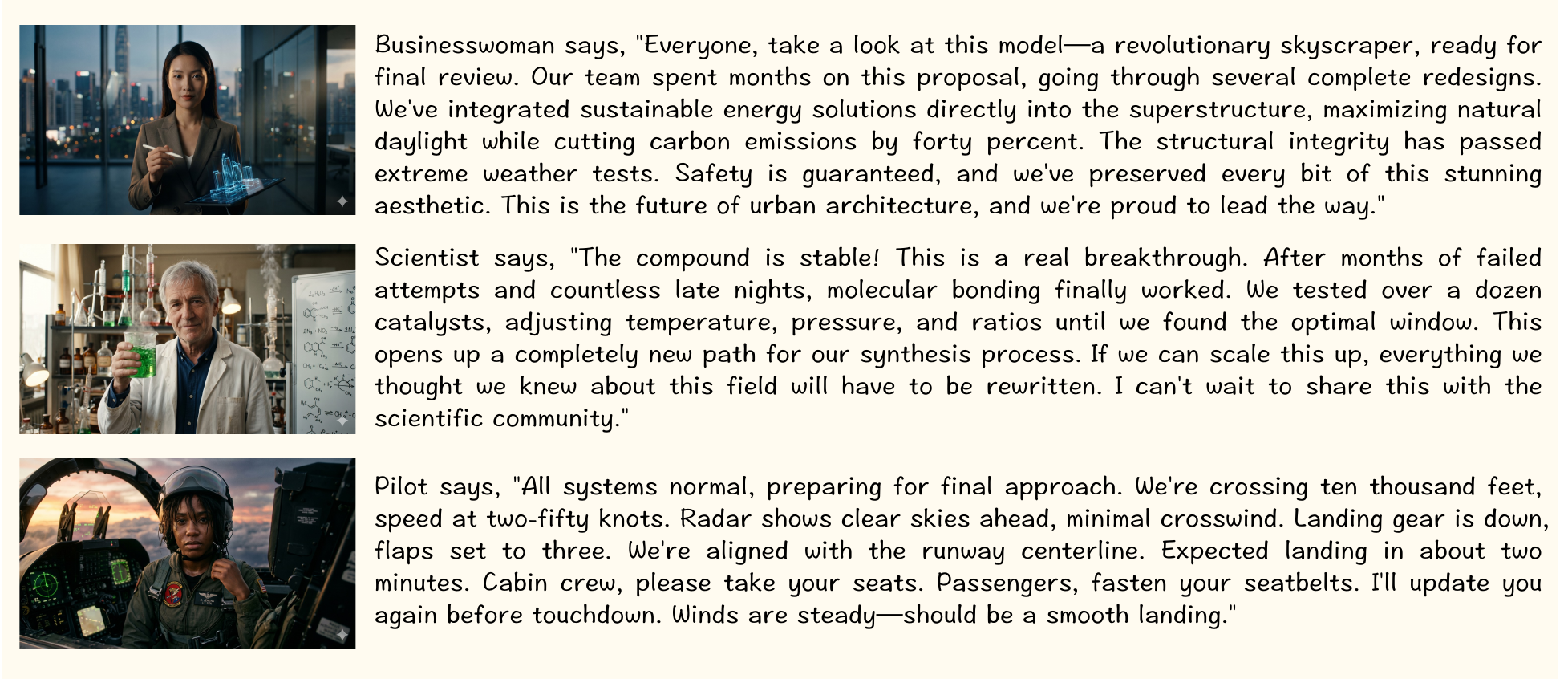}
\caption{\textbf{Examples from our Long-Video Benchmark.} Each example consists of a prompt and its corresponding reference image, where the prompt includes the subject name and speech content for 30-second joint audio-video generation.}
\label{fig:long-video-bench-examples}
\end{figure*}

To better evaluate long-form joint audio-video generation, we construct a long-video benchmark consisting of 50 prompts (25 Chinese and 25 English) with corresponding reference images generated by Gemini~\cite{comanici2025gemini}. All methods are conditioned on the same prompt-image pairs and generate 30-second videos at a resolution of $384 \times 672$. We provide several example pairs in Figure~\ref{fig:long-video-bench-examples}.
Since no ground-truth videos are available, we adopt reference-free evaluation metrics covering audio-visual quality, synchronization, speech alignment, and identity consistency. Together, these metrics provide a comprehensive evaluation of the key capabilities required for long-form audio-video generation.

\begin{itemize}
    \item \textbf{AV Quality.} We evaluate video quality using the average HPSv2~\cite{wu2023hps} score of the middle and last frames, and audio quality using DNSMOS~\cite{reddy2021dnsmos}. The final AV Quality score is the sum of the video quality score and the audio quality score. Higher scores  indicate better audio-visual quality.

    \item \textbf{Speech Align.} We measure speech alignment using Whisper~\cite{radford2023robust} to transcribe the generated speech, and compute the Word Error Rate (WER) and Character Error Rate (CER) between the transcription and the reference speech. The final Speech Align score is defined as: $\text{SpeechAlign} = 1 - (0.5 \times \mathrm{WER}+0.5 \times \mathrm{CER})$. Higher scores indicate better speech alignment.

    \item \textbf{ID Consistency.} We extract DINOv2~\cite{oquab2023dinov2} features from sliding temporal windows and compute the consistency score using the mean feature similarity and the corresponding standard deviation (i.e., $mean - 0.5 \times std$). Higher scores indicate better identity preservation.

    \item \textbf{AV Synchronization.} We evaluate audio-visual synchronization using Synchformer~\cite{iashin2024synchformer} by computing the window-based cosine distance between audio and visual features extracted by the pretrained encoder. The final synchronization score is obtained by applying an exponential transformation $\exp(-x)$ to the distance. Higher scores indicate better synchronization.
\end{itemize}

\section{Additional Ablations}

To investigate the effectiveness of the reinforcement learning stage and the necessity of its key components, we conduct ablation studies from two perspectives: (1) whether the two-phase reward schedule is necessary, and (2) whether the KL penalty is essential for training stability.

\paragraph{Effect of Two-Phase Reward Schedule.}
In Stage 3, we adopt a two-phase reward schedule: we first optimize speech alignment and audio-visual synchronization for 200 steps, then jointly optimize audio-visual quality with the composite reward for another 200 steps. To validate this design choice, we compare our two-phase strategy against a one-stage baseline where all four rewards (speech alignment, audio-visual synchronization, audio quality, and video quality) are combined from the very beginning using the same weights. As shown in Table~\ref{tab:ablation_rl_stage3}, the one-stage baseline results in a notable performance drop, particularly in AV Quality (0.723 vs. 0.751) and AV Synchronize (0.546 vs. 0.579). We hypothesize that jointly optimizing perceptual quality and alignment rewards from the start leads to conflicting gradients, making the policy struggle to balance multiple objectives. The two-phase schedule mitigates this issue by first establishing cross-modal alignment and then refining perceptual quality, ensuring more stable and effective reinforcement learning.

\paragraph{Effect of KL Penalty.}
The KL penalty is designed to constrain the policy update from deviating excessively from the pretrained reference model, thereby preventing reward hacking and ensuring training stability. As shown in Table~\ref{tab:ablation_rl_stage3}, removing the KL penalty leads to a substantial performance drop across all metrics. For example, AV Quality decreases from 0.751 to 0.703. These results demonstrate that without the KL constraint, the model tends to over-optimize the reward signals, resulting in degraded generation quality and poor cross-modal alignment. This confirms that the KL constraint is essential for stable RL post-training under our streaming audio-video generation framework.

\begin{table}[t]
\centering
\begin{tabular}{lccc}
\toprule
\textbf{Component} & \multicolumn{3}{c}{\textbf{Choice}} \\
\midrule
Two-Phase Reward Schedule & {\ding{52}} & \textcolor[RGB]{192, 192, 192}{\ding{56}} & {\ding{52}} \\
KL Penalty & {\ding{52}} & {\ding{52}} & \textcolor[RGB]{192, 192, 192}{\ding{56}} \\
\midrule
{AV Quality$\uparrow$} & \textbf{0.751} & 0.723 & 0.703 \\
{Speech Align$\uparrow$} & \textbf{0.405} & 0.382 & 0.359 \\
{ID Consistency$\uparrow$} & \textbf{0.944} & 0.928 & 0.920 \\
{AV Synchronize$\uparrow$} & \textbf{0.579} & 0.546 & 0.532 \\
\bottomrule
\end{tabular}
\caption{\textbf{Ablation on Components of Stage 3.} We compare our two-phase reward schedule against a one-stage baseline where all rewards are combined from the start, and ablate the KL penalty. Both components are crucial for improvement.}
\label{tab:ablation_rl_stage3}
\end{table}

\section{Human Evaluation}

To complement our automatic metrics, we conduct a user study to assess the perceptual quality of Ripple against existing streaming audio-video generation methods, including Hallo-Live~\cite{li2026hallo} and OmniForcing~\cite{su2026omniforcing}. We sample 25 videos from the VerseBench~\cite{wang2025universe}, and invite 30 participants to choose the best overall model among the three, considering audio-visual quality, synchronization, and speech alignment. As shown in Table~\ref{tab:user_study}, Ripple achieves the highest preference rate, confirming that our automatic evaluation aligns well with human perception.

\begin{table}[t]
\centering
\begin{tabular}{lc}
\toprule
\textbf{Model} & \textbf{Preference Rate (\%)} \\
\midrule
OmniForcing~\cite{su2026omniforcing} & 20.36 \\
Hallo-Live~\cite{li2026hallo} & 30.53 \\
\rowcolor{gray!20}
\textbf{Ripple (Ours)} & \textbf{49.11} \\
\bottomrule
\end{tabular}
\caption{\textbf{Human Preference Rates.} Percentage of participants who preferred each model as the best overall.}
\label{tab:user_study}
\end{table}

\section{Limitation and Discussion}
Despite the promising performance of Ripple, our system still exhibits limited motion diversity in long-form generation. The first-block sink cache, while effective for preserving global scene layout and subject identity across long sequences, introduces a temporal anchoring effect that constrains subsequent blocks from deviating significantly from the initial state, leading to reduced motion magnitude. A promising direction for future work is to decouple identity and timbre preservation from motion dynamics. For example, pre-training the model with explicit identity and timbre conditions could weaken the reliance on the first-block anchor, allowing greater motion variation in subsequent generations.

Beyond this technical limitation, we also consider the potential societal impacts associated with audio-video generation technologies. The capabilities of Ripple may be misused for identity impersonation, unauthorized digital representations, or the creation of deceptive media. Similar to other generative models, our approach may face risks related to deepfakes and misinformation. This work is developed for academic research purposes and does not intend to facilitate misleading content creation or the unauthorized use of personal identities. We emphasize the importance of protecting personal information and obtaining necessary consent. Future work should investigate effective safeguards, such as detection mechanisms and verification strategies, to mitigate potential misuse of audio-video generation systems.

\section{Evaluation Metrics on VerseBench}

VerseBench~\cite{wang2025universe} is a widely used benchmark for evaluating joint audio-video generation. We adopt the evaluation codebase from MOVA~\cite{team2026mova}, which provides a comprehensive evaluation suite covering multiple aspects, including audio-speech alignment, audio-video alignment, and lip synchronization. In this section, we introduce the metrics used to evaluate generated audio and video.

\subsection{Audio-Speech Metrics}

\paragraph{Inception Score (IS).}
We evaluate the quality and diversity of generated audio using the Inception Score~\cite{salimans2016improved}. Specifically, we use a pretrained audio classifier CNN14~\cite{kong2020panns} to extract class probability distributions for each audio sample. The IS is computed as:
\begin{equation}
\mathrm{IS}
=
\exp(\mathbb{E}_x[D_{KL}(p(y|x)\|p(y))]),
\end{equation}
where $p(y|x)$ denotes the conditional label distribution and $p(y)$ represents the marginal distribution. Higher scores indicate better audio diversity and class discriminability.

\paragraph{CLAP.}
We measure the semantic alignment between generated audio and text descriptions using CLAP~\cite{elizalde2023clap}. Specifically, we compute the cosine similarity between the corresponding audio and text embeddings:

\begin{equation}
\mathrm{CLAP}
=
\frac{z_a^Tz_t}{||z_a||||z_t||},
\end{equation}
where $z_a$ and $z_t$ denote the audio and text embeddings extracted by CLAP, respectively. Higher scores indicate stronger audio-text semantic alignment.

\paragraph{DNSMOS.}
We use DNSMOS~\cite{reddy2021dnsmos} to evaluate perceptual audio quality. It is a non-intrusive metric based on subjective ratings and provides three scores: OVRL (overall quality), SIG (signal quality), and BAK (background noise quality). We report the OVRL score, where higher values indicate better audio quality.

\subsection{Audio-Video Metrics}

\paragraph{AV-Align.}
We use AV-Align~\cite{yariv2023diverse} to evaluate temporal synchronization between audio and visual modalities by detecting audio onset peaks and visual motion peaks based on optical flow. The alignment score is computed using the intersection-over-union (IoU) within a temporal window of $1/fps$. Higher scores indicate better synchronization.

\paragraph{IB-Score.}
We evaluate semantic consistency between generated audio and video using ImageBind~\cite{girdhar2023imagebind}. Specifically, we extract audio and video embeddings from generated samples and compute their cosine similarity:
\begin{equation}
\mathrm{IB\text{-}Score}
=
\frac{z_a^Tz_v}{||z_a||||z_v||},
\end{equation}
where $z_a$ and $z_v$ denote the audio and video embeddings extracted by ImageBind, respectively. Higher scores indicate stronger cross-modal semantic consistency.

\subsection{Lip Synchronization Metrics}

\paragraph{LSE-D and LSE-C.}
We use SyncNet~\cite{chung2016out} to evaluate fine-grained lip synchronization between generated speech and facial movements. Specifically, we detect face regions from generated videos and feed the cropped face sequences together with the generated audio into SyncNet. LSE-D~\cite{prajwal2020lip} measures the distance between audio and visual embeddings, where lower values indicate better synchronization. LSE-C~\cite{prajwal2020lip} measures the synchronization confidence, where higher values indicate better lip-audio alignment.

\section{More Visualizations}

\begin{figure*}[!t]
\centering
\includegraphics[width=0.99\textwidth]{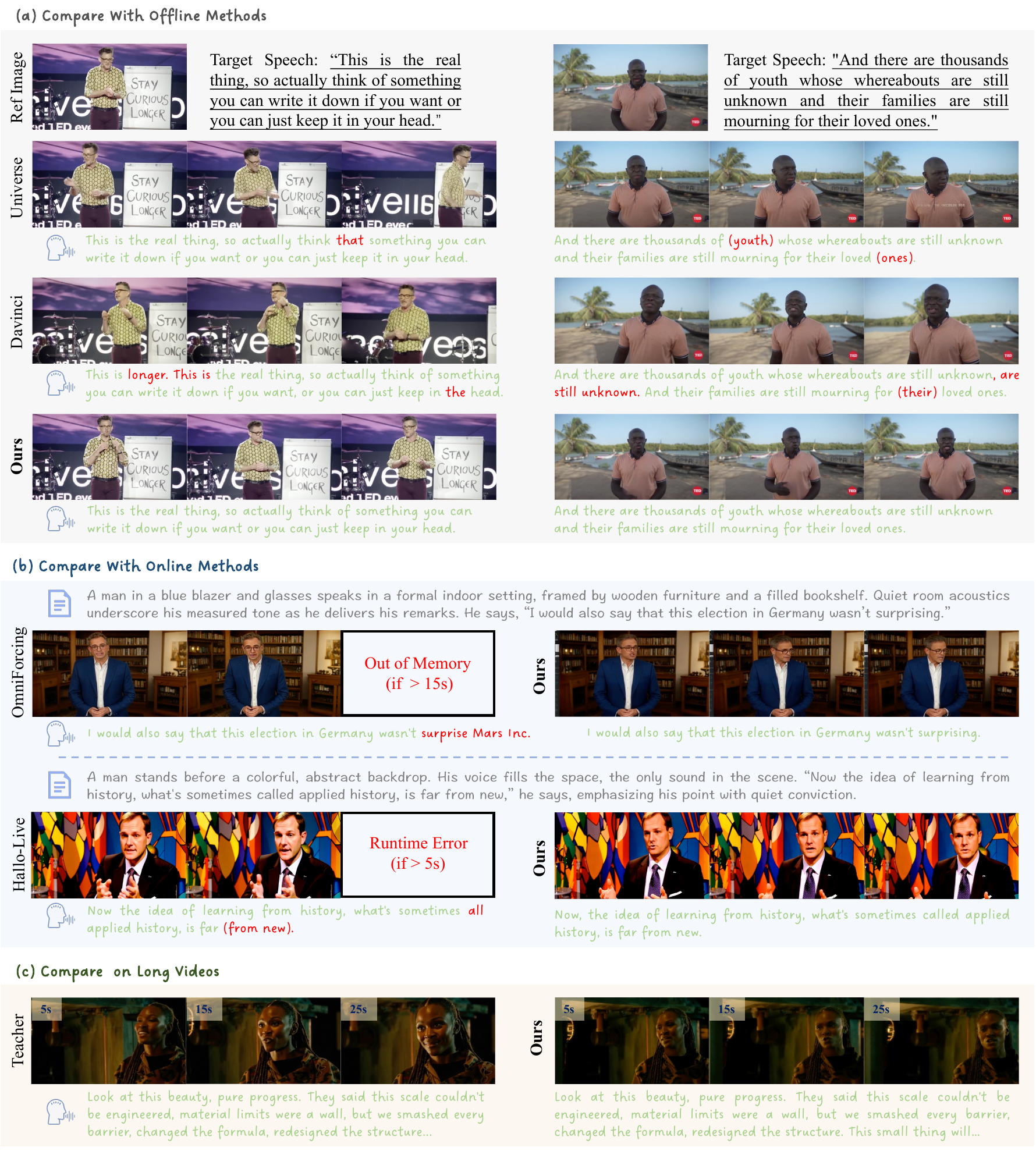}
\caption{\textbf{Qualitative Comparison.} Compared with existing offline and online streaming audio-video generation methods, Ripple achieves high-quality generation on short videos while enabling real-time streaming inference. On long videos, Ripple achieves better temporal consistency than the teacher, demonstrating the effectiveness of our memory mechanism. In the transcribed speech, \textcolor{red}{red} indicates mispronounced/incorrect words, \textcolor{green!50}{green} indicates correct ones, and (parentheses) denote omissions.}
\label{fig:more_compare}
\end{figure*}

\paragraph{Comparison with Baselines.}
As shown in Figure~\ref{fig:more_compare}, we present additional qualitative comparisons against both online and offline streaming audio-video generation methods, where Ripple consistently outperforms the baselines in terms of speech alignment and temporal coherence. Beyond comparing with existing methods, we also compare our streaming model with the bidirectional teacher on long-video generation. Benefiting from our long-term cross-modal memory, Ripple achieves better temporal consistency than the teacher, while requiring substantially less inference latency.

\begin{figure}[!t]
\centering
\includegraphics[width=0.48\textwidth]{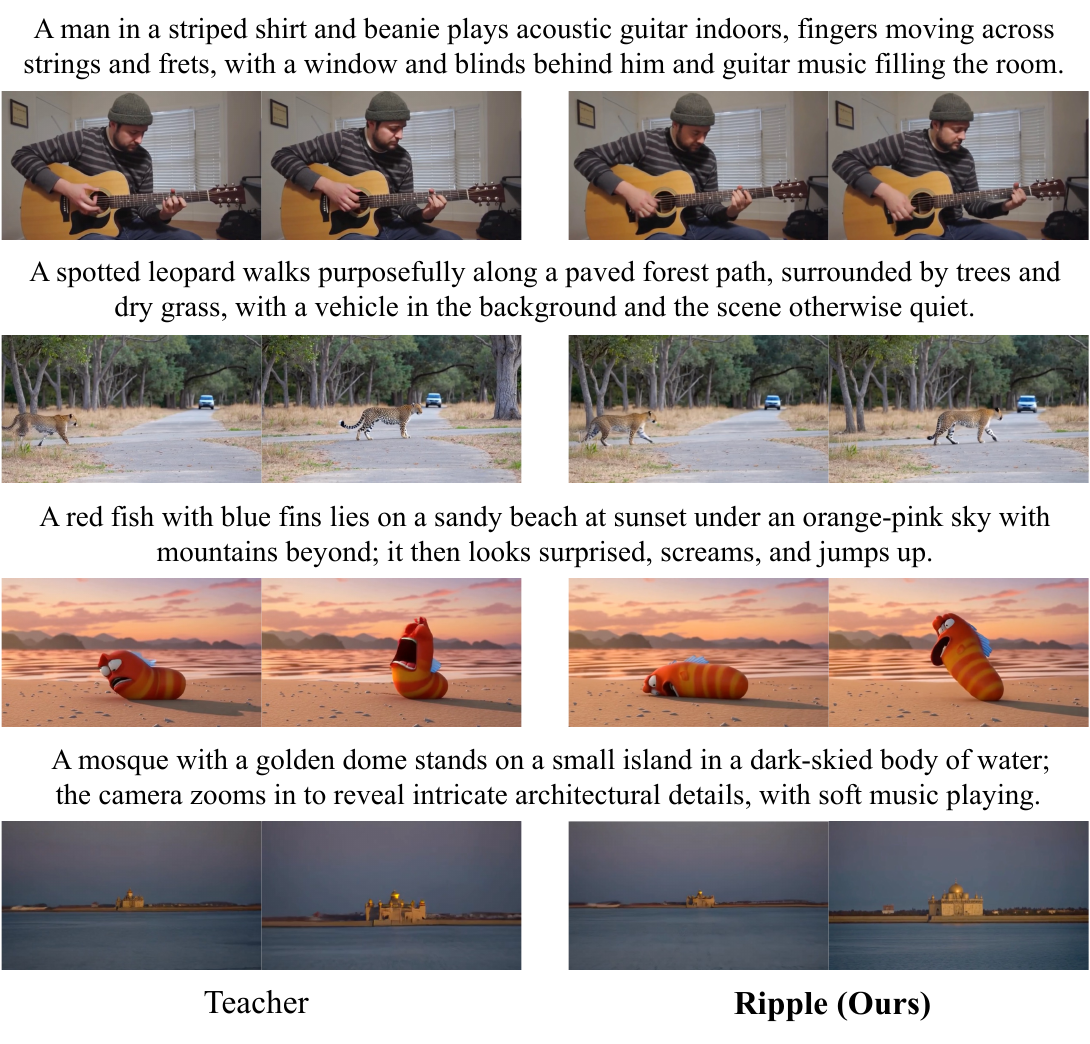}
\caption{\textbf{Comparison on Open-Domain Scenes.} Although trained primarily on human-centered data, Ripple generalizes well to diverse scenarios, achieving generation quality comparable to the teacher while running significantly faster.}
\label{fig:open_domain_compare}
\end{figure}

\paragraph{Open-Domain Generalization.}
Furthermore, we provide additional visualizations on open-domain scenarios in Figure~\ref{fig:open_domain_compare}. Although our training data primarily consists of human-centered content, our method generalizes well to diverse scenarios, including animals, landscapes, and open-world characters, achieving generation quality comparable to the teacher while being significantly faster.

\begin{figure}[!t]
\centering
\includegraphics[width=0.48\textwidth]{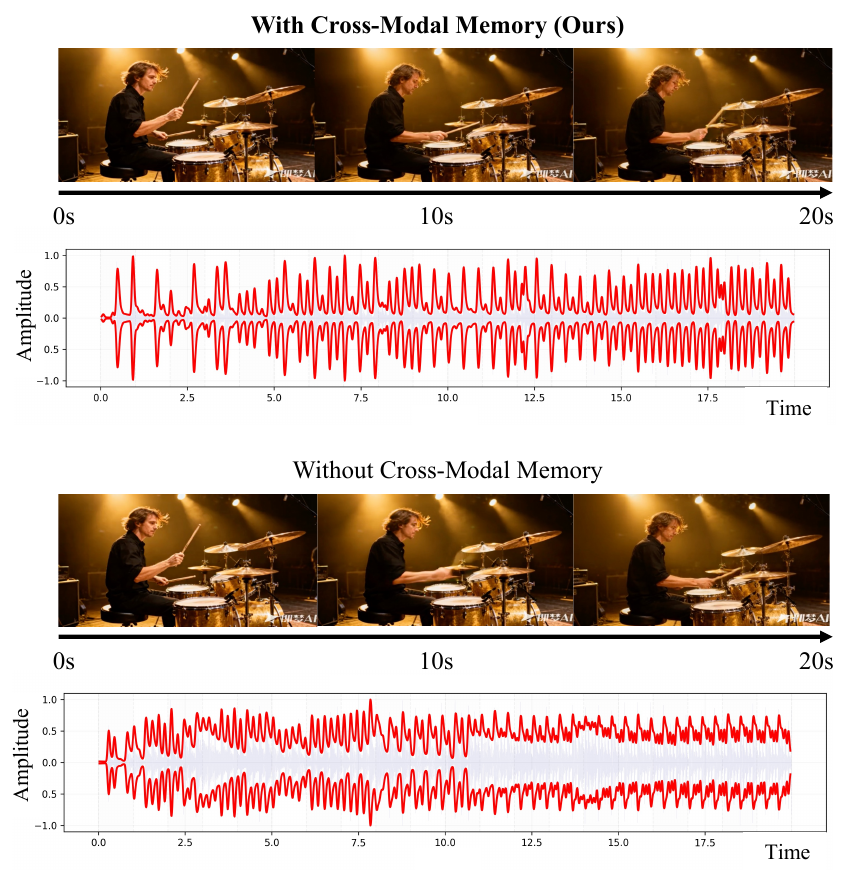}
\caption{\textbf{Audio Waveform Comparison With and Without Cross-Modal Memory.} RMS energy envelopes of generated audio for a drumming scene. With audio memory, the rhythmic energy patterns remain stable and regular throughout the sequence. Without audio memory, noticeable timbre drift and irregular energy fluctuations appear in the later segments (e.g., after 10 s), indicating that the memory mechanism effectively preserves long-term acoustic coherence.}
\label{fig:wave_compare}
\end{figure}

\paragraph{Audio Memory for Timbre Consistency.}
In addition to the visual benefits of our memory mechanism, we provide an audio waveform comparison to demonstrate its role in preserving timbre consistency, as shown in Figure~\ref{fig:wave_compare}. We compare the generated audio for a drumming scene with and without the cross-modal memory. With memory, the rhythmic energy patterns remain regular and stable across the entire sequence, while without memory, the audio exhibits noticeable timbre drift and irregular energy fluctuations in the later segments, indicating that the audio memory effectively maintains long-term acoustic coherence.

\begin{figure}[!t]
\centering
\includegraphics[width=0.48\textwidth]{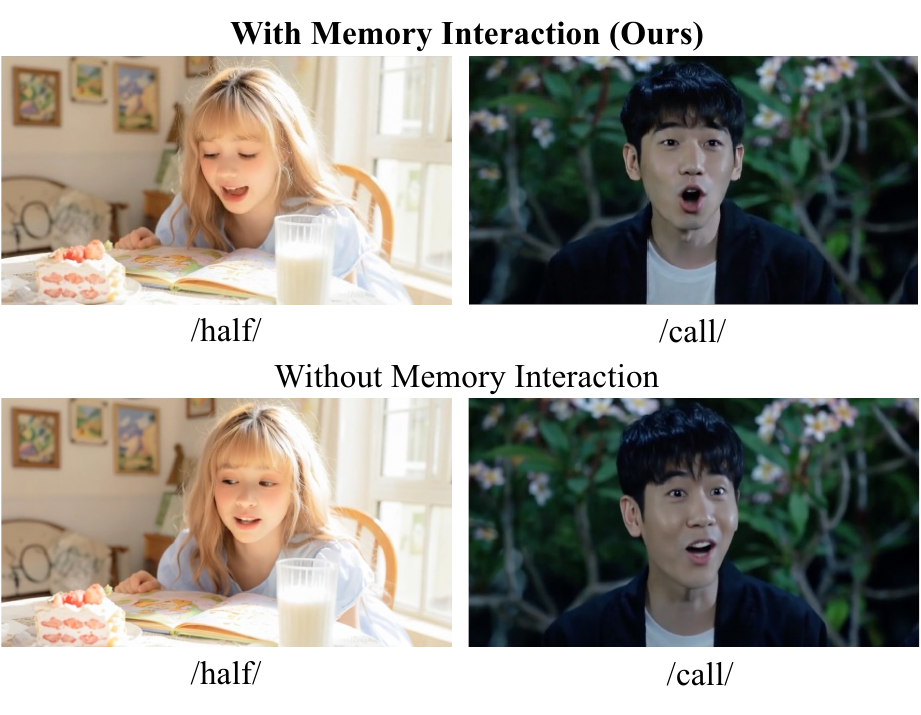}
\caption{\textbf{Qualitative Comparison With and Without Memory Interaction.} The memory interaction module slightly improves audio-visual synchronization, leading to more accurate lip articulation and phoneme pronunciation.}
\label{fig:vis_interaction}
\end{figure}

\paragraph{Effect of Memory Interaction on Lip Synchronization.}
To further demonstrate the effect of memory interaction, we provide a qualitative comparison of lip synchronization in Figure~\ref{fig:vis_interaction}. The memory interaction module brings slight improvement in audio-visual synchronization, leading to more accurate lip articulation, e.g., clearer pronunciation of phonemes such as "/half/" and "/call/". This aligns with our quantitative results, where the full model with memory interaction achieves better AV Synchronize score.

\bibliography{aaai2027}


\end{document}